\documentclass[twoside]{article}
\usepackage{amsfonts}
\usepackage{graphicx}
\usepackage{booktabs}
 \usepackage{tikz}
 \usepackage{hyperref}

 \usetikzlibrary{positioning}

\newcommand{\mig}{\mbox{MIG-TF}}

\newcommand{\lp}{\mbox{TPTF}}
\newcommand{\tuc}{\mbox{TuckER}}

\newcommand{\slava}[1]{\textcolor{black}{#1}}

\definecolor{myred}{RGB}{255,128,128}
\definecolor{myblue}{RGB}{87,133,235}

\usepackage[accepted]{aistats2025}
\usepackage[round]{natbib}

\begin{document}

%

%

\twocolumn[

\aistatstitle{ Knowledge Graph Completion with Mixed Geometry Tensor Factorization}

\aistatsauthor{ Viacheslav Yusupov \And Maxim Rakhuba \And  Evgeny Frolov }

\aistatsaddress{ HSE University \And  HSE University \And  AIRI  \\ HSE University } ]

\begin{abstract}
  In this paper, we propose a new geometric approach for knowledge graph completion via low rank tensor approximation. We augment a pretrained and well-established Euclidean model based on a Tucker tensor decomposition with a novel hyperbolic interaction term. This correction enables more nuanced capturing of distributional properties in data better aligned with real-world knowledge graphs. By combining two geometries together, our approach improves expressivity of the resulting model achieving new state-of-the-art link prediction accuracy with a significantly lower number of parameters compared to the previous Euclidean and hyperbolic models.
\end{abstract}

\section{INTRODUCTION}
Most of the information in the world can be expressed in terms of entities and the relationships between them. This information is effectively represented in the form of a knowledge graph  \citep{kg,kg2}, which serves as a repository for storing various forms of relational data with their interconnections. Particular examples include storing user profiles on social networking platforms \citep{kg-soc}, organizing Internet resources and the links between them, constructing knowledge bases that capture user preferences to enhance the functionality of recommender systems \citep{kg-rec2,llm3}. With the recent emergence of large language models (LLM), knowledge graphs have become an essential tool for improving the consistency and trustworthiness of linguistic tasks. Among notable examples of their application are fact checking \citep{llm1}, hallucinations mitigation \citep{halluc}, retrieval-augmented generation \citep{rag}, and generation of corpus for LLM pretraining \citep{llm2}. This utilization underscores the versatility and utility of knowledge graphs in managing complex datasets and facilitating the manipulation of interconnected information in various domains and downstream tasks.

On the other hand, knowledge graphs may present an incomplete view of the world. Relations can evolve and change over time, be subject to errors, processing limitations, and gaps in available information. Therefore, the tasks of restoring missing links or predicting the new ones are particularly important to maintain the accuracy and relevance of knowledge graphs. Algorithms designed to address this challenge can significantly reduce the manual labor involved in updating information and ensure consistent representation of the most current and up-to-date knowledge. In this work, we aim to design a new efficient graph completion algorithm that will \emph{capture inherent structural properties of the underlying data to improve the link prediction quality}.

The key challenge of this task lies in the choice of proper architectural components of the solution suitable for handling a non-trivial nature of data.
We start by noticing that knowledge graphs exhibit a strictly non-uniform arrangement of interconnections. Some nodes on the graph are present more frequently than others, which renders a hierarchical structure often reproducing the features of a power law-like distribution. It places knowledge graphs into the so called ``complex network'' category. In the seminal work on the theoretical foundations of complex networks analysis \citep{kruk}, it was shown that hyperbolic geometries are especially suitable for modelling this type of structures.
In these geometries, as one moves further away from the origin, the distances between points and the area grow exponentially in contrast to the Euclidean geometry with its respective linear and quadratic growth. Due to this effect, hyperbolic embeddings have a higher expressive ability in terms of capturing hierarchical relations within data.
Consequent studies have proved the practicality of such geometric approach in various domains such as natural language processing \citep{nick}, computer vision \citep{hypvit2022} and recommender systems \citep{hypvae2020}.

\begin{figure}[h]
\centering
\includegraphics[width=0.5\textwidth]{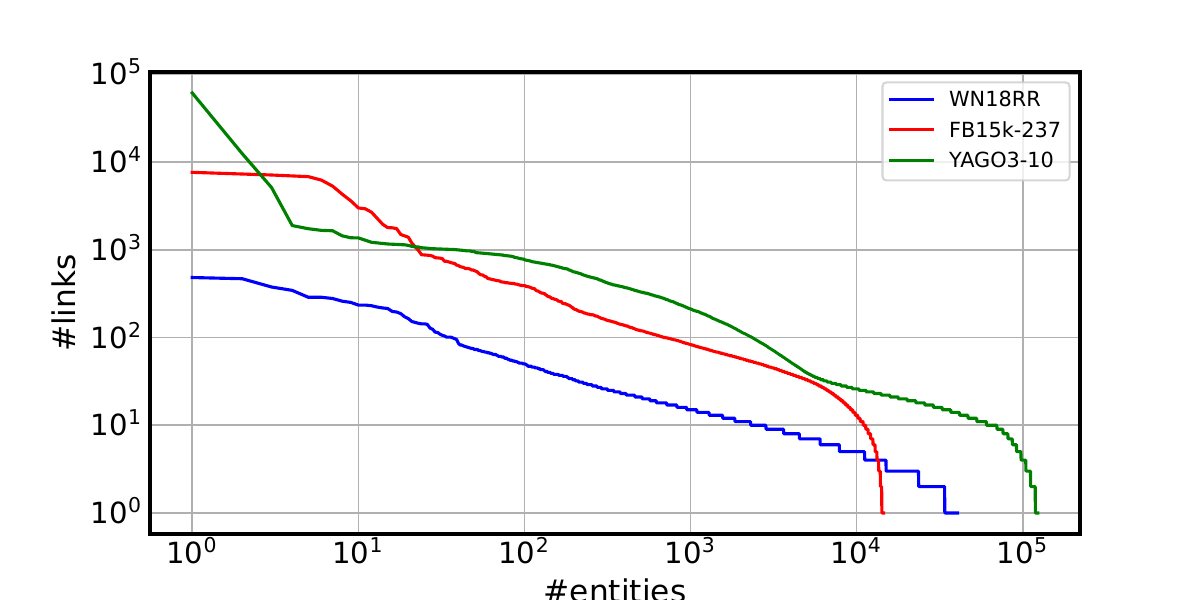}
\caption{Links distribution on three benchmark knowledge graphs considered in this work. The FB15k-237 dataset violates the power law the most, which is indicated by the longest platue in the leftmost part of the curve. Correspondingly, the top-performing model on this dataset among previous state-of-the-art turns out to be Euclidean. In contrast, our \mig{} approach outperforms both Euclidean and hyperbolic models, see Table~\ref{table:results}.}
\label{fig:powerlaw}
\end{figure}

\begin{figure*}[ht]
	\centering
	\includegraphics[width=1\textwidth]{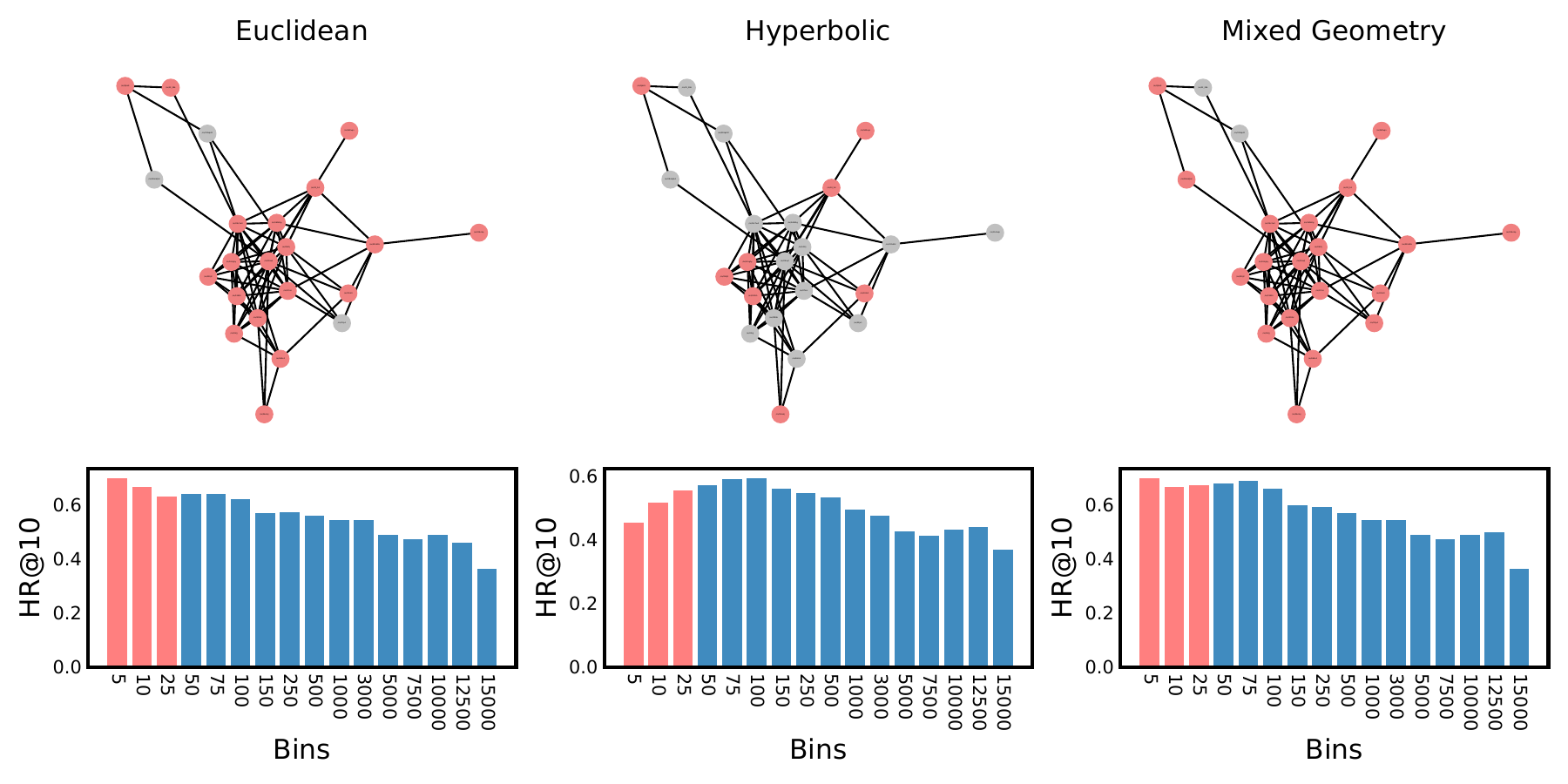}
 
	\caption{The graphs at the top row demonstrate the top 20 vertices in terms of their number of connections in the FB15k-237 knowledge graph. The pink marks indicate vertices that were predicted in more than $50\%$ of cases by both the Euclidean (left), hyperbolic (center) and mixed geometry (right)  models. The bar charts illustrate the hit rate (HR@10) of the predictions for links between vertices from each group. Vertices are arranged in descending order based on the number of incoming links. As can be observed, due to the high number of relations among active vertices, the hyperbolic model (\lp{}  ) significantly underperforms in predicting relations between active vertices compared to the Euclidean model(\tuc{}). However, the combination of Euclidean and hyperbolic models outperforms both of them. }
	\label{fig:topvertices}
\end{figure*}

Despite the promising results, in our experiments, we observe that real-world knowledge graphs may not consistently align with the assumption of strict hierarchical internal structure and may only partially follow a power-law distribution (see Figure ~\ref{fig:powerlaw}). More specifically, the characteristics of relations in a knowledge graph or even data preprocessing may render connectivity components consisting of vertices with a large number of links almost uniformly distributed between them. We will call these vertices ``active''. As we demonstrate in our work, \emph{having a large number of active vertices violates hierarchy and disrupts the total compatibility of hyperbolic geometry with real data} (see Figure ~\ref{fig:topvertices}).
Therefore, in this paper, \emph{we aim to find a balanced solution using different geometry types} so that the weakness of each geometry is compensated. We introduce a hybrid mixed geometry model that combines the strengths of both Euclidean and hyperbolic geometry within a single architecture. In this model, the Euclidean component serves the goal of properly extracting information from the active vertices, while the hyperbolic component with embeddings from a hyperbolic space captures additional information from vertices in the graph that are distributed according to a power-law.

Overall, our contributions can be summarized as follows:
\begin{itemize}

\item We introduce a new mixed-geometry tensor factorisation (\mig{}) model that combines Tucker decomposition defined in the Euclidean space with a new low-parametric hyperbolic ternary interaction term.
\item We highlight intricacies of applying geometric approach to real-world knowledge graphs and demonstrate the associated with it limitations of using single-geometry modelling.
\item The proposed combined approach significantly reduces the number of model parameters compared to state-of-the-art methods. It does so without sacrificing expressive power and achieves more accurate results in most of the common benchmarks. 
\end{itemize}

The rest of the paper is organized as follows.
In Section~\ref{sec:related}, we review most relevant Euclidean and hyperbolic models.
We introduce our parameter-efficient hyperbolic interaction model tetrahedron pooling tensor factorization (\lp{})   along with the basics of hyperbolic geometry in Section~\ref{sec:lpitf}.
Section~\ref{sec:migtf} is devoted to mixed geometry modelling \mig{}, where we present the mixture of the Euclidean \tuc{} and the hyperbolic \lp{}   models.
Finally, Section ~\ref{sec:experiments} contains the results of numerical experiments, where we showcase the performance of our model and also investigate the impact of using a hyperbolic correction. The anonymized code for reproducing our results is publicly available\footnote{https://github.com/hse-cs/MIGTF}.

\section{RELATED WORK} \label{sec:related}
Knowledge graphs represent facts in the triplet form $(e_s, r, e_o)$, where $e_s$ and $e_o$ denote a pair of subject and objects entities and $r$ denotes the corresponding relation between them. This representation can be naturally expressed in the third-order binary tensor form where a non-zero entry at position $(i, j, k)$ encodes an existing $(e_i, r_k, e_j)$ triplet of the knowledge graph. To identify connections in knowledge graphs that have been incorrectly specified in the interaction tensor, various tensor factorization-based models are utilized.

For instance, the \emph{SimplIE} \citep{simple} and \emph{ComplEX}\citep{Complex}  models employ Canonical decomposition ($CP$) \citep{canon} with embeddings in the real or complex Euclidean space. The \emph{PITF} \citep{pitf} model utilizes Pairwise Interaction Tensor Factorisation in the Euclidean space, which splits the standard ternary interaction term of $CP$ into a linear sum of independent interactions between entity-entity and entity-relation pairs. Some other models use distances between embeddings of entities and relations in the vector space. For instance, \emph{TransE} \citep{transe} uses Euclidean vector representations of relations, which are used as connection vectors between Euclidean embeddings of entities. The further development of this approach named \emph{Quate} \citep{quate} represents embeddings as rotations in the hypercomplex space. 

Various artificial neural networks-based approaches were also proposed to solve  the link prediction task. For instance, the models like \emph{ConvE} and \emph{ConvKB} \citep{conv, conv3}  utilize convolutional neural networks to work with knowledge graphs. The \emph{R-GCN} \citep{3gcn} model proposes a modification of graph-convolutional neural networks \emph{GCN} \citep{gcn} to adapt graphs convolutions to relational data. Models, such as \emph{GAAT} \citep{gaat} or \emph{KBGAT} \citep{kbgat}, use the attention mechanisms on graphs.

Some models rely on the \emph{shared-factors paradigm}, which utilizes a special structure of the problem based on the fact that objects and subjects belong to the same space. While making the problem inherently non-linear, this approach significantly reduces the number of learned parameters of models and is better suited for predicting symmetric links \citep{TuckER} in knowledge graphs. For instance, the \emph{MEKER} \citep{meker}) model employs \emph{CP} decomposition with low-rank shared factors, and \emph{RESCAL} \citep{rescal} utilizes \emph{DEDICOM} \citep{dedicomp} factorisation decomposing each tensor slice along the relation axis. The \tuc{} \citep{TuckER}  and \emph{R-TuckER} \citep{reim} models employ shared-factors \tuc{}  decomposition \citep{tuc} to represent knowledge graph entities and relations and supports the prediction of symmetric as well as asymmetric links in knowledge graphs.

Recent advances in understanding of hyperbolic geometry applications has lead to the development of hyperbolic models for knowledge graph completion as well. According to \citep{kruk}, the structural properties of hyperbolic spaces make them plausible for analyzing a special family of non-uniformly distributed data, which knowledge graphs also belong to. For instance, the \emph{RotH} and \emph{RefH} \citep{Roth} propose specific rotation and reflection transformations in a hyperbolic space to improve the expressive power of the learned embeddings. Even though these model show remarkable improvements on some knowledge graphs, \emph{there are still some cases where Euclidean models outperform hyperbolic ones}. We attribute this observation to the violation of some basic distributional properties required for full compatibility with hyperbolic geometry. It motivates us to develop a hybrid approach similarly to \citep{gu2019, hybrid1, hybrid2} that takes the best from both geometries and therefore compensates for their shortcomings in analyzing real-world data. \slava{ Similar idea was utilized in models \emph{M$^2$GNN} \citep{mgnn} generalizing graph neural network for multi-curvature and \emph{GIE} \citep{gie} introducing message passing and attention algorithms on hyperbolic and hyperspherical spaces. In comparison to previous methods, our \mig{} approach uses less complex and more stable operations (Section \ref{sec:comp}). Additionally, it implies learning only low-parametric hyperbolic addition to a pretrained model instead of learning the whole model as in \emph{RotH}, \emph{M$^2$GNN} and \emph{GIE}. } Further in the text, we provide the detailed description of our approach.

\section{MIXED GEOMETRY TENSOR FACTORIZATION}

In this section, we introduce our new parameter-efficient hyperbolic interaction term \lp{}, and the new mixed geometry model \mig{}, where the hyperbolic term is used as a correction to the Euclidean model.

\subsection{Hyperbolic geometry and Lorentz Distance}
We start with basic notation and concepts from hyperbolic geometry.
There are several models of hyperbolic geometry, among which Poincare Ball, Klein, and Lorentz model are frequently used in various machine learning approaches. All hyperbolic models are isomorphic -- they describe the same geometry and can be interchangeably utilized by means of the corresponding transformations \citep{ungar}. In our work, we use the Lorentz model of hyperbolic geometry. In the following, we only introduce the basics of Lorentz geometry and refer the reader to \citep{lgeom} for more details. 

Let us define the Lorentz inner product for vectors $x, y \in \mathbb {R}^{n + 1}$, which is used for measuring distance between hyperbolic embeddings:

\[
\langle x, y \rangle_{\mathcal{L}} = -x_0y_0 + \sum_{i = 1}^{n} x_i y_i, \label{scal_prod}
\]
where
\[
x_0 = \sqrt{\beta + \sum_{i = 1}^{n} x_i^2}. 
\]

The induced norm of a vector $x \in \mathbb {R}^{n + 1}$ in Lorentz geometry is defined as $\|x\|_{\mathcal{L}}^2 = \langle x, x \rangle_{\mathcal{L}} = -\beta$.
The corresponding $n$-dimensional space $\mathcal{H}^{n, \beta} \subset \mathbb {R}^{n + 1}$ is called Hyperboloid and defined as follows:
\[
\mathcal{H}^{n, \beta} = \left\{x \in \mathbb {R}^{n + 1}\,\big|\ \|x\|_{\mathcal{L}}^2 = -\beta, \ \beta \ge 0\right\}.\label{hyp_boloid}
\]
The origin vector of the hyperboloid $\mathcal{H}^{n, \beta}$ equals to $\textbf{0} = (\beta, 0, ..., 0) \in \mathbb {R}^{n + 1}$. The inner product of $\textbf{0}$ and $x$ is, hence, $\langle \textbf{0}, x \rangle_{\mathcal{L}} = -\beta x_0$.

The associated geodesic distance is defined as 
\[
d_l(x, y) = arccosh(-\langle x, y \rangle_{\mathcal{L}}).
\]

Similarly to \citep{lfm}, we introduce the square Lorentz distance between $x, y \in \mathcal{H}^{n}$ is defined as 
\begin{equation}
    d^2_{\mathcal{L}}(x, y) = \|x - y\|_{\mathcal{L}}^2 = -2 - 2\langle x, y \rangle_{\mathcal{L}}. \label{dist}
\end{equation}
This distance fulfills all of the Euclidean distance axioms, except for the triangle inequality:
\begin{equation} 
d(x, z) \le d(x, y) + d(y, z).
\label{eq:triang}
\end{equation}
Nevertheless, it does not hold in general in the hyperbolic setting. 
The idea to benefit from this fact appears, for instance, in a two-dimensional hyperbolic model \emph{LorentzFM} \citep{lfm}. Inequality violation in this model implies that two entities are far away and, hence, become irrelevant to each other. 
Generalizing the triangle inequality to the three-dimensional case is not a straightforward task, which we address in the following section.

\subsection{Tetrahedron Pooling Tensor Factorization}
\label{sec:lpitf}
Solving the knowledge graph link prediction problem requires learning ternary interactions in the form of entity-relation-entity triplets. This can be naturally modelled through tensor representation in the Euclidean space. However, unlike the 2D case, generalizing a representation of such third-order interactions to the hyperbolic space is not straightforward. To the best of our knowledge, there is no unified approach.

We propose to modify~\eqref{eq:triang} to capture ternary interactions. In particular, we utilize the so-called tetrahedron inequality \citep{tetraed}: for the points $u,v,t,o$ in the Euclidean space, it holds
\begin{equation}\label{eq:tetraineq}
 d(u, v) + d(o, t) \le d(u, t) + d(v, t) + d(o, u) + d(o, v).
\end{equation}
In the hyperbolic context, this inequality does not universally apply. However, as we explore further, the regions where this inequality is violated can be advantageous for our objectives. Consequently, we can naturally introduce the following score function:

\begin{equation}\label{eq:scorelor}
\begin{aligned}
S_G = &d_l(u, v) + d_l(\textbf{0}, t) -d_l(u, t)\\
 &- d_l(v, t) - d_l(\textbf{0}, u) - d_l(\textbf{0}, v),
\end{aligned}
\end{equation}
where $u, v \in \mathbb {R}^{ d_l}$ are  entity embeddings and $t \in \mathbb {R}^{ d_l}$ is the relation embedding. 
Note that this score function is negative when inequality~\eqref{eq:tetraineq} holds and is positive in the opposite case.
Unfortunately, the score function~\eqref{eq:scorelor} involves arccosh, which is not differentiable everywhere and, hence, may lead to difficulties for gradient-based optimization methods.

\begin{figure}[t]
	\centering
	\includegraphics[width=0.45\textwidth]{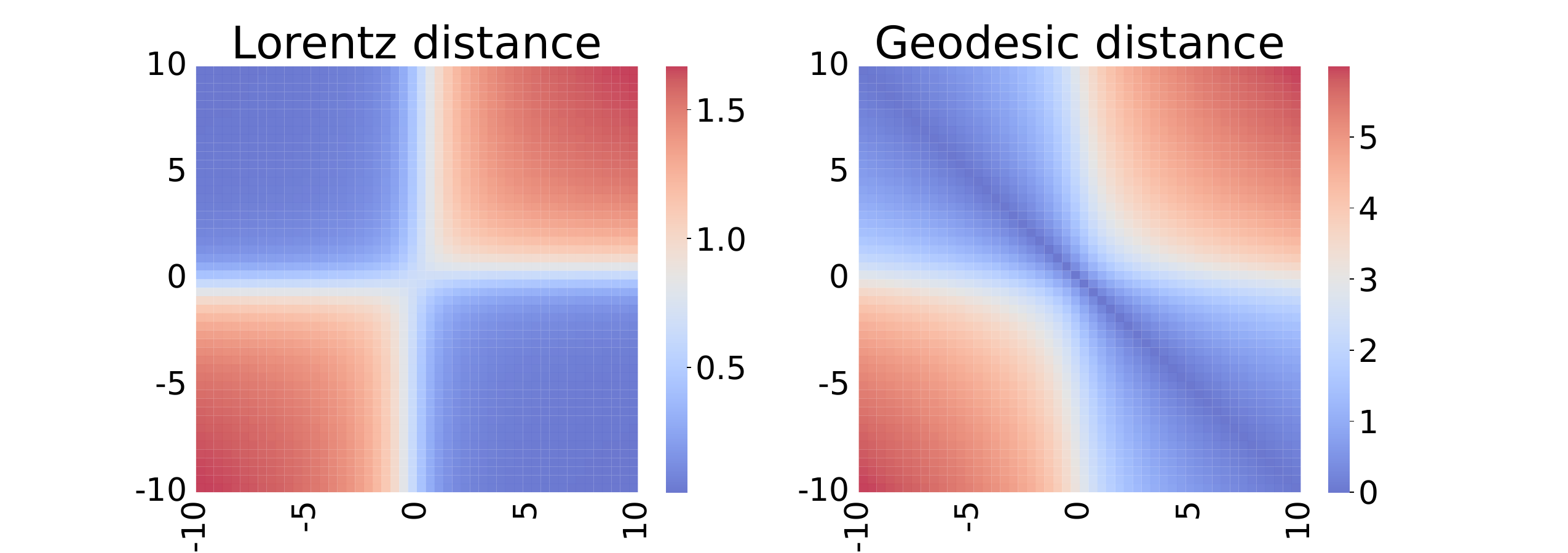}
 
	\caption{Left score function landscape corresponds to our score function \eqref{lpitf_scor}, whilst right one corresponds to~\eqref{eq:scorelor}. As seen from the plots the landscapes of score functions for different distances are similar. For more score function landscapes see Appendix.}
	\label{fig:landscapes}
    \end{figure}

Similarly to \emph{LorentzFM}, \eqref{eq:scorelor} is replaced with the following ``smoothed'' score function that mimics the desired behaviour:
\begin{equation}
\begin{aligned}
&S_H(u, v, t) = \\
&\frac{\dfrac{1}{2} \left(\begin{array}{c}d_\mathcal{L}^2(u, v) + d_\mathcal{L}^2(\textbf{0}, t) - d_\mathcal{L}^2(u, t) \\ \  
   - d_\mathcal{L}^2(t, v) - d_\mathcal{L}^2( \textbf{0}, u)
  - d_\mathcal{L}^2(\textbf{0}, v) \end{array} \right)}{\langle \textbf{0}, v \rangle_{\mathcal{L}}\langle \textbf{0}, t \rangle_{\mathcal{L}} + \langle \textbf{0}, u \rangle_{\mathcal{L}}\langle \textbf{0}, v \rangle_{\mathcal{L}} + \langle \textbf{0}, t \rangle_{\mathcal{L}}\langle \textbf{0}, v \rangle_{\mathcal{L}}},
\end{aligned}
\label{lpitf_scor}
\end{equation}
see Figure \ref{fig:landscapes} for comparison of the two score functions.

Note also that we may rewrite $S_H$ as:
\[
S_H =  \frac{(2 + t_0 -u_0 - v_0) - (\langle u, v \rangle_\mathcal{L} - \langle u, t \rangle_\mathcal{L} - \langle t, v \rangle_\mathcal{L})}{u_0v_0 + u_0t_0 + t_0v_0}.
\]

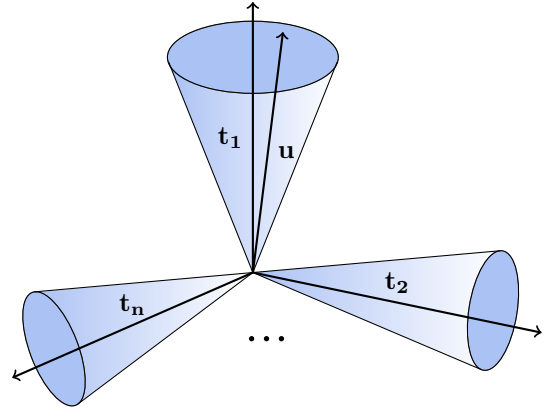
\begin{figure}
\centering
 \begin{tikzpicture}[scale=0.8]
 
  \def\x{1.4}
  \def\y{3.5}
  \def\R{\x+0.02}
  \def\yc{\y+0.08}
  \def\e{0.6}
  \def\op{0.5}
  \def\opp{1.5}
 
  \shade[right color=white,left color=myblue,opacity=\op]
    (-\x,\y) -- (-\x,\yc) arc (180:360:{\R} and \e) -- (\x,\y) -- (0,0) -- cycle;
  \draw[fill=myblue,opacity=\op]
    (0,\yc) circle ({\R} and \e);
  \draw
    (-\x,\y) -- (0,0) -- (\x,\y);
  \draw
    (0,\yc) circle ({\R} and \e);
 
  \def\x{1}
  \def\y{3.5}
  \def\R{\x+0.005}
  \def\yc{\y+0.04}
  \def\e{0.4}
 
  \begin{scope}[rotate=111]
    \shade[right color=white,left color=myblue,opacity=\op]
      (-\x,\yc) -- (-\x,\yc) arc (180:360:{\R} and \e) -- (\x,\yc) -- (0,0) -- cycle;
    \draw[fill=myblue,opacity=\op]
      (0,\yc) circle ({\R} and \e);
    \draw
      (-\x,\y) -- (0,0) -- (\x,\y);
    \draw
      (0,\yc) circle ({\R} and \e);
  \end{scope}

  \def\x{1.0}
  \def\y{4.0}
  \def\R{\x+0.005}
  \def\yc{\y+0.04}
  \def\e{0.4}
 
  \begin{scope}[rotate=-99]
    \shade[right color=white,left color=myblue,opacity=\op]
      (-\x,\yc) -- (-\x,\yc) arc (180:360:{\R} and \e) -- (\x,\yc) -- (0,0) -- cycle;
    \draw[fill=myblue,opacity=\op]
      (0,\yc) circle ({\R} and \e);
    \draw
      (-\x,\y) -- (0,0) -- (\x,\y);
    \draw
      (0,\yc) circle ({\R} and \e);
  \end{scope}

  \draw[->, thick, black] (0,0) -- (0.5, 4) node[midway, right] {\(\mathbf{u}\)};
  \draw[->, thick, black] (0,0) -- (0, 4.5) node[midway, left] {\(\mathbf{t_1}\)};
  \draw[->, thick, black] (0,0) -- (4.8, -1.0) node[midway, above] {\(\mathbf{t_2}\)};
  \draw[->, thick, black] (0,0) -- (-4.0, -1.75) node[midway, above] {\(\mathbf{t_n}\)};
  \node[black!80!black,right] at (-0.3, -1.1) {\huge{$...$}};
 
\end{tikzpicture}
\label{cones}
\caption{Each embedding of a relation ($t_1, ..., t_n$) defines a cone that encompasses the embeddings of the entities associated with that relation.}
\label{fig:cones}
\end{figure}

The score function yields a positive value when the points corresponding to linked entities are in close proximity and a negative value when they are distant (see Appendix). Additionally, we note that the terms $d_\mathcal{L}^2(u, t)$ and $d_\mathcal{L}^2(t, v)$, which account for interactions between relations and entities, effectively draw entities $u$ and $v$ closer to the relation $t$. Consequently, the trained model exhibits a distinct behavior: entities tend to cluster around well-separated relations, as illustrated in Figure~\ref{fig:cones}.

The model's predictions~$p$ for entity $u$ and relation $t$ can be obtained with $p = \{p_j\}_{j=1}^{n_e}$:
\[
p = \sigma(S_H(u, V, t)),\] 
where $V \in \mathbb {R}^{d_l \times n_e}$ is the matrix of all entity embeddings and $n_e$ -- the number of entities. In \lp{}  model, we use the binary cross-entropy (BCE) loss function:
\[
l_{BCE}(y_i, p_j) = - y_i \log(p_j) - (1 - y_i) \log(1 - p_j),
\]
 where $y  = \{y_i\}_{i=1}^{n_e} $ -- binary label vector of the ground truth.
 Similarly to \citep{lorentz2}, we optimize $u$, $v$ and $t$ embeddings in the Euclidean space with the \emph{AdamW} \citep{adamw} optimizer and then map them to the hyperboloid as follows: $v \in \mathbb {R}^{n}$ maps to  $[\sqrt{\beta + \|v\|_2^2}, v] \in \mathbb {R}^{n + 1} $.

\subsection{Mixed Geometry Tensor Factorisation}
\label{sec:migtf}
\begin{figure}
\centering
\begin{tikzpicture}[auto, scale=1, every node/.style={scale=0.85}]

\tikzstyle{block} = [rectangle, draw, text centered, minimum height=2em, minimum width=10em]
\tikzstyle{block2} = [rectangle, draw, text centered, minimum height=2em, minimum width=4em]
\tikzstyle{frozen} = [block, fill=myblue!75]
\tikzstyle{trainable} = [block, fill=myred!65]
\tikzstyle{sum} = [circle, draw, text centered, minimum size=3em]
\tikzstyle{arrow} = [draw, -latex]

\node [frozen] (tucker) {TuckER (frozen)};
\node [trainable, below=1cm of tucker] (lpitf) {TPTF (trainable)};
\node [sum, below right=0.17cm and 0.4cm of tucker] (sum) {\textbf{+}};
\node [sum, right=0.5cm of sum] (sigma) {$\sigma$};
\node [block2, right=0.5cm of sigma] (loss) {Loss};
\node [block, below=0.75cm of lpitf] (update) {Update TPTF weights};

\draw[arrow] (tucker) -| (sum);
\draw[arrow] (lpitf) -| (sum);
\draw[arrow] (sum) -- (sigma);
\draw[arrow] (sigma) -- (loss);
\draw[arrow] (loss)  |- (update);
\draw[arrow] (update) -- (lpitf);

\draw[dashed] ([xshift=-0.5cm, yshift=0.5cm]tucker.north west)
  rectangle ([xshift=2.62cm, yshift=-0.5cm]lpitf.south east);
\end{tikzpicture}
\caption{The proposed \mig{} model architecture.}
\label{scheme}
\end{figure}
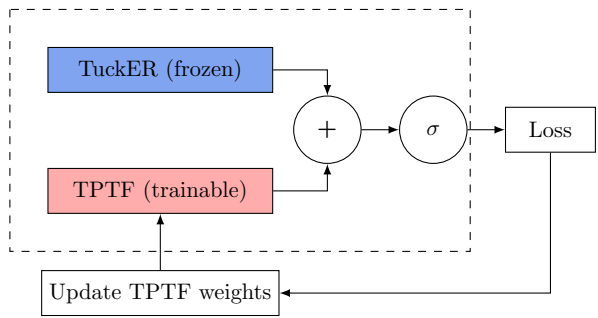

To improve the link prediction in knowledge graphs we introduce a shared-factor mixed geometry model combining Euclidean and hyperbolic models (see Figure \ref{scheme}). Due to the complexity of the knowledge graph, some vertices are better described with hyperbolic embeddings whilst others are with the Euclidean ones. Therefore with a combination of hyperbolic and Euclidean embeddings, significantly more information on the structure of the knowledge graph is saved with Euclidean and hyperbolic vector representations together. For the hyperbolic model, we used \lp{} model \eqref{lpitf_scor}, and for Euclidean one we used \tuc{}   \citep{TuckER}. The score function of the Euclidean model is defined as:

\begin{equation}
S_E(G, u, v, t) = \sum_{\alpha, \beta, \gamma = 1}^{d_1, d_1, d_2} G_{\alpha \beta \gamma}  u_{\alpha}  v_{\beta}  t_{\gamma}, \label{tucker} 
\end{equation}
 where $u, v \in \mathbb {R}^{ d_1}$ -- embeddings of entities, $t \in \mathbb {R}^{d_2}$ -- embedding of relations and $G \in \mathbb {R}^{d_1 \times d_1 \times d_2}$ --  tensor of learnable parameters. Here, $d_1$ and $d_2$ are the ranks of the Tucker decomposition.

 The score function of our mixed-geometry model \mig{} is the sum of score functions of the Lorentzian and Euclidean models: 
\begin{equation}
    (S_{\textit{MIG-TF}})_i = (S_E)_i + (S_H)_i, \label{mig_tf}
\end{equation}
where $(S_E)_i = S_E(G, u, V_i, t)$ defined in \eqref{tucker} is an Euclidean interaction term,  $(S_H)_i = S_H(u^{\mathcal{L}},V_{i}^{\mathcal{L}},t^{\mathcal{L}})$ is a hyperbolic interaction term \eqref{lpitf_scor}. The vectors $u$ and $t$ Euclidean entity and relation embeddings, $u^{\mathcal{L}}$ and $t^{\mathcal{L}}$ -- hyperbolic embeddings,  $V_i$ and $V_i^{\mathcal{L}}$ are $i$-th rows of matrices $V$ and $V^{\mathcal{L}}$ of $n_e$  Euclidean and hyperbolic entity embeddings, respectively,  and $G \in \mathbb {R}^{d_1 \times d_1 \times d_2}$ -- 3-D tensor of learnable parameters. In \mig{} model, we utilize pretrained \tuc{} model \eqref{tucker} and optimize the hyperbolic term parameters \eqref{lpitf_scor} of the score function~\eqref{mig_tf} to minimize the BCE loss:
\[
\mathcal{L}_{\textit{MIG-TF}} = \frac{1}{n_e} \sum_{i = 1}^{n_e} l_{BCE}(a_{i}, \,\sigma((S_{\textit{MIG-TF}})_i)),\label{optim}
\]
where $a_{i}$ is an $i$-th element of interaction vector $a$ for entity and relation with embeddings $u$ and $t$, respectively. We optimize hyperbolic embeddings as in Section \ref{sec:lpitf}. To prevent model overfitting, we employ the dropout layer \citep{dropout} in both Euclidean and hyperbolic interaction terms.

\subsection{\slava{Mixed-geometry models comparison}} \label{sec:comp} 
\slava{The next subsection is devoted to comparing our approach with other mixed-geometry methods: \emph{M$^2$GNN} \citep{mgnn} and \emph{GIE} \citep{gie}.}

The translational distance \citep{trdist} model \emph{M$^2$GNN} generalizes graph neural networks for multi-relational setting on knowledge graphs and modified it for multi-curvature setting capturing graph structures more effectively, especially hierarchical and cyclic ones.

The model \emph{GIE} introduces the message passing and attention  mechanism for the knowledge graph completion by utilizing mapping on hyperbolic and hyperspherical spaces.

\slava{Both of the existing approaches suffers from computational complexity and numerical instability due to utilizing exponential and logarithmic mapping, which are even more numerically unstable than the standard Lorentz model operations \citep{hdnn}. In contrast, \mig{} proposes using Square Lorentz distance, a convenient replacement for Geodesic distance. Square Lorentz distance reduces hyperbolic operations, such as exponential and logarithmic mappings and arccosh, to mere inner product computations, decreases complexity of operations and increases numerical stability.}

\slava{Additionally, only the low-parametric addition \lp{} is learned in our \mig{} method instead of learning the whole models in \emph{M$^2$GNN} and \emph{GIE}.}

\slava{Unlike \citep{mgnn} and \citep{gie}, we do not utilize spherical geometry and still our mixed geometry approach achieves the best metrics in the majority of cases. Nevertheless, our framework is sufficiently flexible to accommodate the incorporation of spherical geometry, presenting a promising direction for future research.}

\section{EXPERIMENTS}
\label{sec:experiments}
In experiments, we shows that mixed geometry model, combining Euclidean and hyperbolic embeddings, with low-dimensional hyperbolic ones outperforms Euclidean, Complex and hyperbolic models with significantly higher dimensions of embeddings. Additionally,  we have studied the dependence of curvature of the space of hyperbolic embeddings and expressive ability of hyperbolic and mixed geometry models. Moreover, we have shown that hyperbolic addition to \tuc{} \citep{TuckER} model improves the quality of link prediction between hierarchically distributed entities.

\subsection{Datasets} 
We evaluate our mixed geometry model for link prediction task on three standard knowledge graphs: WN18RR \citep{transe, conv}, FB15k-237 \citep{transe, fb15k} and YAGO3-10 \citep{yago3-10}. WN18RR is a subset of WordNet \citep{wordnet} -- knowledge base containing lexical relations between words. FB15k-237 is a subset of FreeBase \citep{fb} -- collaborative knowledge base containing general world knowledge, such as nationalities of famous people, locations of some buildings or genre of music. YAGO3-10 is a subset of YAGO3 \citep{yago} which mostly contains descriptions o people. The statistics of these knowledge graphs are shown in following table:
\begin{table}[ht]
\caption{Knowledge graphs statistics.}
\label{table:1}
\footnotesize
\centering
\scalebox{1}{
\begin{tabular}{lccc}
\toprule
$\mathsf{Dataset}$ & $\mathsf{Entities}$ & $\mathsf{Relations}$ & $\mathsf{Triplets}$ \\
\midrule
$WN18RR$ & $40,943$& $11$ & $93,003$ \\
$FB15k\text{-}237$ & $14,541$& $237$ & $310,116$ \\
$YAGO3\text{-}10$ & $123,182$& $37$ & $1,179,040$ \\
\bottomrule
\end{tabular}}
\end{table}
For each knowledge graph we follow standard data augmentation by adding inverse relations \citep{Bordes}.

\subsection{Baselines}
We compare our models with $8$ baselines: $4$ models with euclidean embeddings, $2$ with complex embeddings and $2$ with hyperbolic embeddings:
\begin{itemize}
\item \textbf{Euclidean models}: \emph{PITF} (Pairwise Interaction Tensor Factorisation) -- model using PITF tensor decomposition and Euclidean embeddings \citep{pitf}. \emph{DistMult} (Distance-based Multiplication) -- model utilizing canonical tensor decomposition \citep{DistMult}.
\item \textbf{TuckER-based Euclidean models}: \tuc{} (Tucker decomposition) -- model using Tucker tensor decomposition and Euclidean embeddings \citep{TuckER}.
\emph{RTuckER} -- \tuc{} model with a Riemanian method of optimization \citep{reim}.
\item \textbf{Complex models}: \emph{ComplEx-N3} (Complex Embeddings) -- model using real part of the canonical tensor decomposition and complex embeddings \citep{ComplexN3}. \emph{RotatE} (Relational Rotation in complex space) -- model using rotations in complex space and complex embeddings \citep{Rotate}.
\item \textbf{Hyperbolic models}: \emph{RotH} (Hyperbolic Rotations) and \emph{RefH} (Hyperbolic Reflections) -- attention models using rotations and reflections in hyperbolic space and utilizing hyperbolic embeddings from Poincare Ball \citep{Roth}.
\item \textbf{Mixed-geometry models}: \emph{M$^2$GNN} (Mixed-Curvature Multi-Relational Graph Neural Network) -- model using multi-curvature embeddings for translational distance and multi-relational graph neural networks. \emph{GIE} (Geometry Interaction Knowledge Graph Embeddings) -- model utilizing attention layers and embeddings from Euclidean, hyperbolic and spherical spaces.
\end{itemize}

To demonstrate the benefits of our mixed geometry approach to link prediction task we compare baselines with our hyperbolic model \lp{}  \eqref{lpitf_scor} and our mixed geometry models \mig{} \eqref{mig_tf} and \mig{}$_{QR}$. Here, \mig{}$_{QR}$ is mixed geometry model with orthogonalization heuristic. For more details see Appendix.  Additionaly, to better understand the role of hyperbolic interaction term we compare \lp{}  and \mig{} models with different fixed hyperbolicities.

\subsection{Evaluation metrics} 
At the test mode, we rank correct entities against all entities using score fictions \eqref{lpitf_scor} for \lp{}  and \eqref{mig_tf} for \mig{}. We compare models using two rank-base metrics: $MRR$ (Mean  Reciprocal Rank) and $HR@k,$ $k \in {1, 3, 10}$ (Hit Rates). $MRR$ measures mean inverse rank of correct entity, whilst $HR@k$ measures the share of correct triples among $Top-k$ recommendations \citep{metr}. We follow the standard evaluation protocol in the filtered setting: all true triples in the KG are filtered out during evaluation \citep{dat-aug}.
 
\section{RESULTS}\label{sec:results} 
\begin{figure}[ht]
	\centering
	\includegraphics[width=0.47\textwidth]{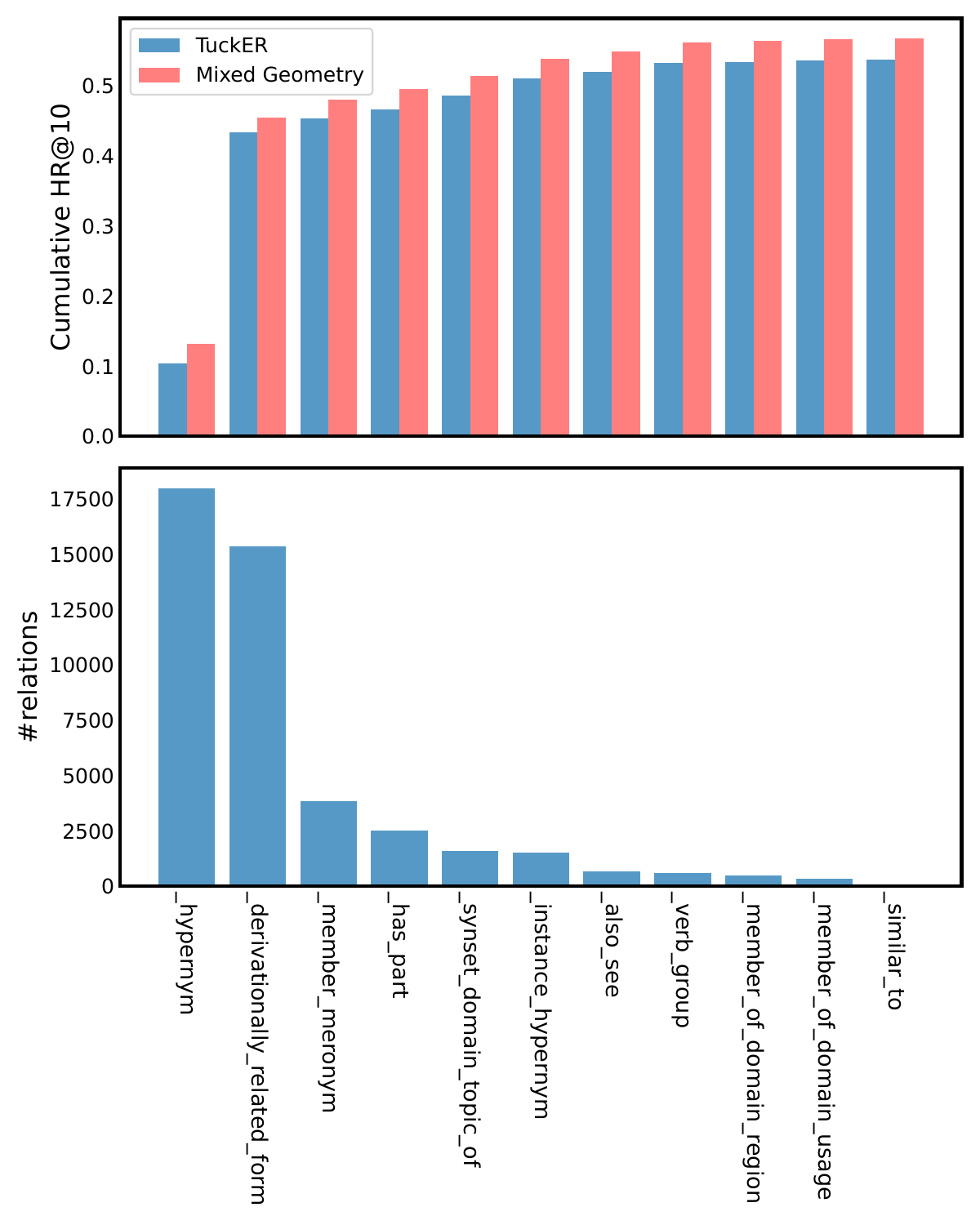}
 
\caption{The top bar chart demonstrates the cumulative $HR@10$ metrics in the $WN18RR$ knowledge graph with respect to all types of relations ordered by their occurrence frequency. It is seen that Mixed Geometry model (red) has higher cumulative metrics in comparison to  \tuc{} (blue). In addition, Mixed Geometry has significant increase in $HR@10$ metric on rare relations. For instance, on  ``member of domain region'' $+140\%$, ``member of domain usage'' $+48\%$ and ``member meronym'' $+36\%$. The bottom bar chart illustrates the number of links with each relation in knowledge graph.}
	\label{fig:6}
    \end{figure}

Our mixed geometry models achieved new state-of-the-art results on FB15k-237, YAGO3-10 and WN18RR in the majority of metrics. Moreover, our models have significantly less parameters then the best models on WN18RR and YAGO3-10. For instance, our models have $8$ times less parameters than \emph{RotH} or \emph{RefH}.  Additionally, as we expected, Lorentz addition to the TuckER model significantly increases $MRR$ and $HR@k$ metrics, with low - Lorentzian embeddings. We also show, that \mig{} model is better that \tuc{} model in prediction of various relations, especially rare ones Figure~\ref{fig:6}. 
\begin{table}[h]
\caption{Approximate number of  models' parameters  on knowledge graphs $WN18RR$, $FB15k-237$ and $YAGO3-10$.}

	\label{table:params}
	\footnotesize
	\centering
        \scalebox{0.9}{
	\begin{tabular}{lccc}
		\toprule
  
		\textsf{Models}  &  $\mathsf{FB15k\text{-}237}$ & $\mathsf{WN18RR}$ & $\mathsf{YAGO3\text{-}10}$ \\
		\midrule
            $PITF$     & $4 \cdot 10^6$ & $8 \cdot 10^6$ & $25 \cdot 10^6$  \\
            $DistMult$  &  $4 \cdot 10^6$ & $8 \cdot 10^6$ & $25 \cdot 10^6$ \\
            $TuckER_{S_E+0\cdot S_H}$ & $4 \cdot 10^6$ & $8 \cdot 10^6$  & $25 \cdot 10^6$\\
            $RTuckER_{S_E+0\cdot S_H}$ & $4 \cdot 10^6$ & $8 \cdot 10^6$  & $25 \cdot 10^6$\\
            $ComplEx\text{-}N3$  &  $4 \cdot 10^6$ & $8 \cdot 10^6$ & $25 \cdot 10^6$\\
            $RotatE$  & $60 \cdot 10^6$ & $60 \cdot 10^6$ & $120 \cdot 10^6$\\
            $RotH$  & $40 \cdot 10^6$ & $80 \cdot 10^6$ & $120 \cdot 10^6$\\
            $RefH$  & $40 \cdot 10^6$ & $80 \cdot 10^6$ & $120 \cdot 10^6$\\
            $M^2GNN$ & $12 \cdot 10^6$ & $24 \cdot 10^6$ & $75 \cdot 10^6$\\
            $GIE$ & $4 \cdot 10^6$ & $18 \cdot 10^6$ & $24 \cdot 10^6$\\
            \midrule
            \textsf{Our models}& \multicolumn{3}{c}{}\\
            \midrule
		$TPTF_{0 \cdot S_E + S_H} $  &   $2 \cdot 10^6$ &   $4 \cdot 10^6$ & $12 \cdot 10^6$\\
            $MIG\text{-}TF_{S_E+S_H}$  & $5 \cdot 10^6$ & $10 \cdot 10^6$ & $31 \cdot 10^6$ \\

		\bottomrule
	\end{tabular}}
        \vspace{0.2cm}

\end{table}

\begin{table*}[ht]
\caption{Metrics on knowledge graphs: $HR@k$ and $MRR$ metrics of models on knowledge graphs WN18RR, FB12k-237 and YAGO3-10. \textbf{Bold} means the best metric on the knowledge graph and \underline{underlined} -- the second best metric. For each knowledge graph we follow standard data augmentation by adding inverse relations \citep{Bordes}.}
	\label{table:results}
	\footnotesize
	\centering
	\scalebox{0.90}{\begin{tabular}{lcccccccccccc}
		\toprule
		& \multicolumn{4}{c}{$\mathsf{FB15k\text{-}237}$} &
		\multicolumn{4}{c}{$\mathsf{WN18RR}$} & \multicolumn{4}{c}{$\mathsf{YAGO3\text{-}10}$} \\ 
            \midrule
  
		\textsf{Models}  &  $\mathsf{HR@10}$ & $\mathsf{HR@3}$ & $\mathsf{HR@1}$ & $\mathsf{MRR}$ &  $\mathsf{HR@10}$ & $\mathsf{HR@3}$ & $\mathsf{HR@1}$ & $\mathsf{MRR}$ & $\mathsf{HR@10}$ & $\mathsf{HR@3}$ & $\mathsf{HR@1}$ & $\mathsf{MRR}$\\
		\midrule
            $PITF$     &  0.459 & 0.347 & 0.186 & 0.284 & 0.403 & 0.384& 0.392 & 0.296 & 0.610 & 0.494 & 0.350 & 0.453\\
            $DistMult$  & 0.419 & 0.263 & 0.155 & 0.241 & 0.490 & 0.440 & 0.390 & 0.430 & 0.540 & 0.380 & 0.240 & 0.340 \\ 
            $TuckER_{S_E + 0\cdot S_H}$  & 0.544 & \underline{0.394} & \underline{0.266} & 0.358 & 0.526 & 0.482 & 0.443 & 0.470 & 0.681 & 0.579 & 0.466 & 0.544 \\
            $RTuckER_{S_E + 0\cdot S_H}$  & 0.505 & 0.387 & 0.237 & 0.326 & 0.546 & 0.495 & 0.446 & 0.479 & -- & -- & -- & -- \\
             $ComplEx\text{-}N3$  & 0.547 & 0.392 & 0.264 & 0.357 & 0.572 & 0.495 & 0.435 & 0.480 & 0.701 & 0.609 & 0.498 & 0.569 \\
            $RotatE$  & 0.533 & 0.375 & 0.245 &0.338 & 0.571 & 0.492 & 0.428 & 0.476 & 0.670 & 0.550 & 0.402 & 0.495 \\
            $RotH$  & 0.535 & 0.380 & 0.246 & 0.344 & \textbf{0.586} & \textbf{0.514} & 0.449 & \underline{0.496} & 0.706& 0.612& 0.495 & 0.573\\
            $RefH$  & 0.536 & 0.383 & 0.252 & 0.346 & 0.568 & 0.485 & 0.404 & 0.461 & 0.711& 0.619& \underline{0.502} & 0.576\\
            $M^2GNN$ & \textbf{0.565} & 0.398 & \underline{0.275} & 0.362 & 0.572 & 0.498 & 0.444 & 0.485 & 0.702 & 0.605 & 0.478 & 0.573 \\
            $GIE$ & 0.552 & \underline{0.401} & 0.271 & 0.362 & \underline{0.575} & 0.505 & \textbf{0.452} & 0.491 & 0.709& 0.618 & \textbf{0.505} & \underline{0.579} \\ 
            \midrule
             \textsf{Our models} & \multicolumn{12}{c}{}\\
            \midrule
            $TPTF_{0\cdot S_E+ S_H}$  & 0.504 & 0.321 & 0.186 & 0.238 & 0.489 & 0.378 & 0.252 & 0.314 & 0.665 & 0.521 & 0.383 & 0.481\\ 
            $MIG\text{-}TF_{S_E+S_H}$  & \underline{0.554} & \textbf{0.402} & \textbf{0.277} & \textbf{0.367} & 0.561 & \underline{0.507} & \underline{0.450} & \underline{0.496} & \underline{0.717} & \underline{0.621} & \underline{0.502} & \underline{0.579} \\
            $MIG\text{-}TF_{QR, S_E+S_H}$  & 0.553 & \textbf{0.402} & \textbf{0.277} & \underline{0.366} & 0.574 & \textbf{0.514} & \textbf{0.452} & \textbf{0.499} & \textbf{0.720} & \textbf{0.622} & \underline{0.502} & \textbf{0.580} \\

		\bottomrule
	\end{tabular}}

\end{table*}
 
\subsection{Hyperbolicity} 
\begin{figure}[ht]
	\centering
	\includegraphics[width=0.5\textwidth]{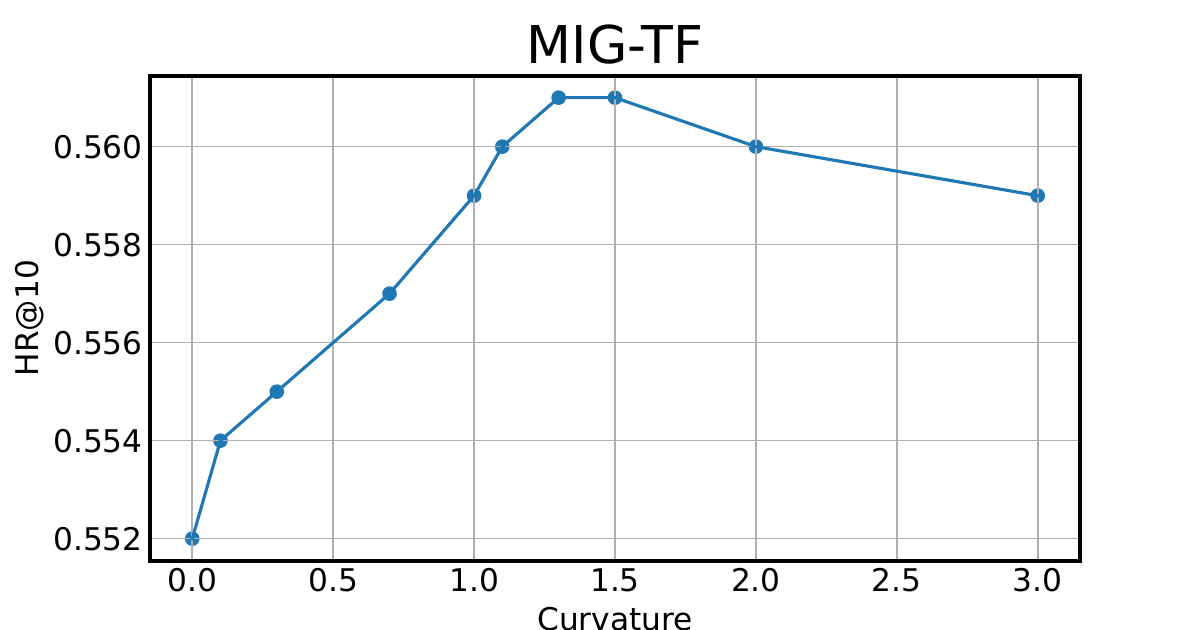}
 
	\caption{The graph presents the dependence of $HR@10$ metric from curvature of Hyperbolic space $\beta$ for \mig{} model on WN18RR dataset. }
	\label{fig:5}
    \end{figure}

To study the impact of hyperbolic interaction term in hybrid model we trained \lp{}  and \mig{} models with various fixed curvatures. Models with low curvature are similar to euclidean ones, therefore they worse predict links between hierarchically distributed entities of knowledge graph. In contrast, models with curvature $c \approx 1.5$  show the highest metrics both in mixed geometry and hyperbolic cases Figure~\ref{fig:5}.

Euclidean component in \mig{} model embeds features of knowledge graph structure, whilst hyperbolic ones are not fully represented with embeddings. Therefore, the hyperbolic component of mixed geometry model represents only the hyperbolic features of knowledge graph that left after Euclidean component application \citep{ni2015, gu2019}. In contrast, \emph{RotH} and \emph{RefH} architectures \citep{Roth}, are based on rotations and reflections in hyperbolic and the Euclidean space, respectively.  In contrast, our model utilizes the hyperbolic triangle inequality, as well as uses embeddings in both Euclidean and hyperbolic spaces.  Additionally, our model has less stringent requirements on the structure of the input data.

\subsection{Link prediction quality}
As demonstrated in Table~\ref{table:results}, our mixed geometry approach with mixed geometries outperforms all other competitors on the two out of three datasets. It slightly underperforms its main competitor (the hyperbolic \emph{RotH} model) on the $WN18RR$ dataset on one metric. We attribute this result to the fact that the structure of this dataset is the most plausible for analysis with hyperbolic geometry. Indeed, by inspecting Figure~\ref{fig:powerlaw}, we see that this dataset is the most compatible with the ideal power law distribution, it contains the minimal amount of active nodes and exhibits almost linear dependency over the majority of entities. This result aligns well with our initial design that is targeted towards mixed hierarchical and non-hierarchical structures and therefore is may not take full advantage of more homogeneous data organization. A possible remedy to this effect would be changing Euclidean component of \mig{} model to more powerful model, such as Euclidean \emph{RotE} \citep{Roth} or complex \emph{ComplEx-N3} \citep{ComplexN3}. We leave investigation of this approach for future work.

On the other two datasets our approach dominates its competitors, which aligns well with our assumptions on the structural complexities of these datasets that prevent purely hyperbolic or purely Euclidean model to learn the best representations. Our approach not only outperforms other models but also requires a lesser amount of parameters, which indicates the higher representation capacity of the mixed geometry embeddings. 
By consulting to Figure~\ref{fig:powerlaw}, one can see that the most prominent results are obtained on the $FB15k-237$ dataset exhibiting the highest degree of violation of the power law distribution. 
Remarkably, the otherwise top-performing hyperbolic models underperform the best Euclidean model on this dataset. This signifies the detrimental effect of the structural non-homogeneity on single-geometry models and highlight the advantages of our mixed geometry approach.

\section{CONCLUSION}
The use of the proposed mixed geometry model \mig{} combining the Euclidean and hyperbolic embedding techniques can significantly improve the quality of link predictions in knowledge graphs with non-homogeneous structure. It provides a significant advantage over state-of-the-art single-geometry models, such as the Euclidean \tuc{} and the hyperbolic \emph{RotH} and \emph{RefH}.
This effect can be attributed to the fact that both Euclidean and Hyperbolic embeddings capture different aspects of relationships between entities in a knowledge graph. By combining these two approaches,  mixed geometry models can capture a broader range of relationships, leading to more accurate predictions. 
We also show that hybrid models are particularly well-suited for addressing the challenge of link prediction in knowledge graphs under limited computing resources. Due to the geometry-driven improved representation capacity, our proposed approach leads to fewer parameters compared to complex or attention-based models such as \emph{RotatE} and \emph{RotH}. \slava{Additionally, our model requires training only the minority of its parameter in comparison to the other mixed geometry models such as \emph{GIE} and \emph{M$^2$GNN}. This facts make \mig{} approach more efficient and scalable.}

\subsubsection*{Acknowledgments}
The work of Viacheslav Yusupov and Maxim Rakhuba was prepared within the framework of the HSE University Basic Research Program. The calculations were performed in part through the computational resources of HPC facilities at HSE University \citep{hsehpc}.

\bibliographystyle{plainnat}
\bibliography{references}

\onecolumn

\section{SQUARE LORENTZ DISTANCE VS. GEODESIC DISTANCE}

\begin{figure}[ht]
	\centering
	\includegraphics[width=1.0\textwidth]{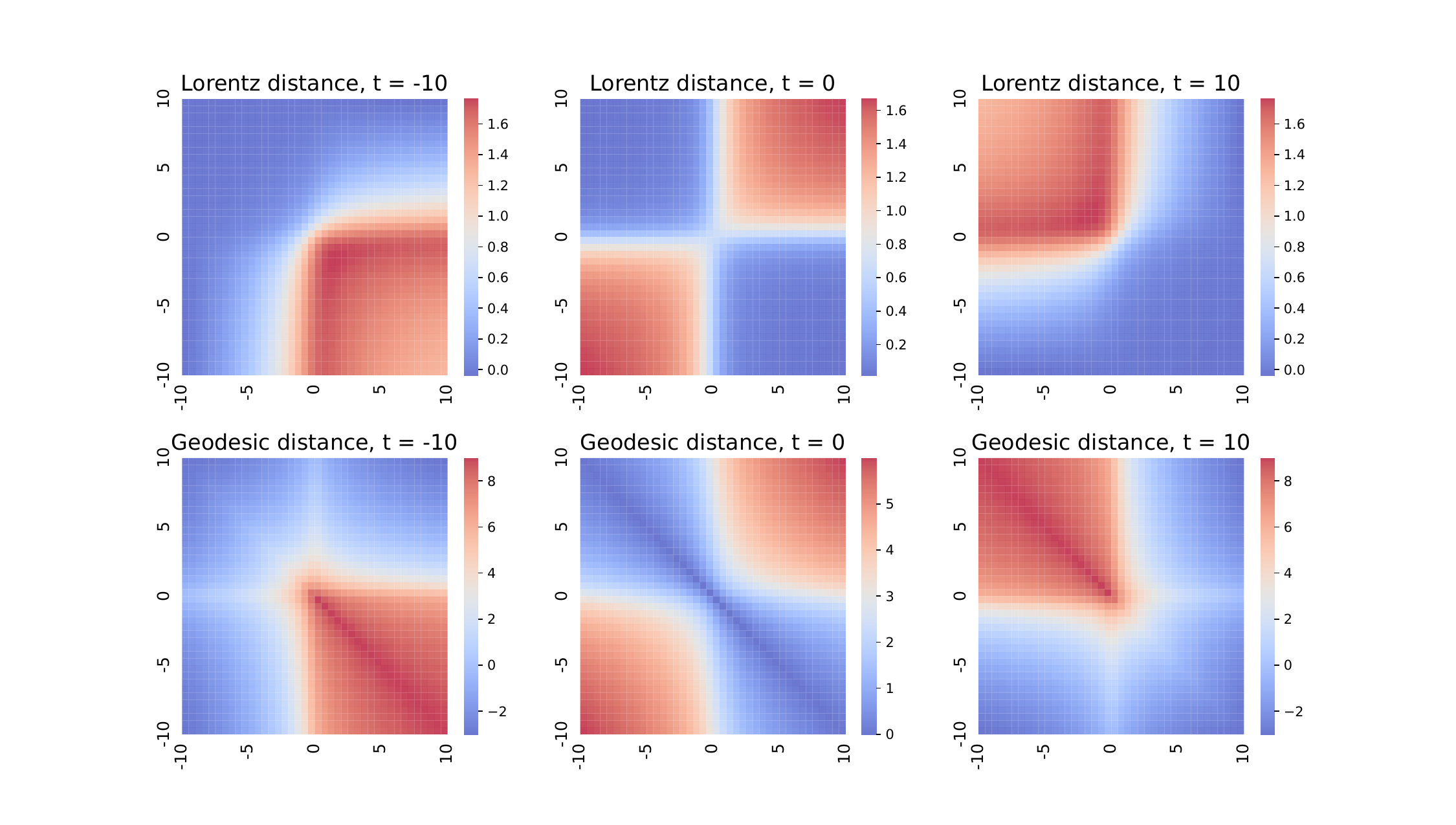}

	\caption{The landscapes of the \lp{} score function for the Lorentz and geodesic distances. As is seen in the plots, the normalized score function with the Lorentz distance mimics the behaviour of ones with geodesic distance.}
	\label{fig:heatmaps_full}
    \end{figure}

In this section, we illustrate that the score function based on the utilized Lorentz distance behaves similarly to the geodesic distance for different $t$. In particular, in Figure~\ref{fig:heatmaps_full}, we present the two-dimensional landscapes of the \lp{} score function  for $t \in \{-10, 0, 10\}$.

\section{COMPLEXITY}
In this section, we estimate the complexity of our \mig{} model.

Let us introduce the operation $G \times_i V$ for a tensor  $G \in \mathbb {R}^{d_1 \times d_2 \times ...\times d_n}$ and matrix $V \in \mathbb {R}^{ d_i \times m}$
\[
(G \times_i V)_{a_1, ...,a_{i - 1}, k, a_{i + 1}, ..., a_n } = \sum_{a_i = 1}^{d_i} G_{a_1, a_2, ..., a_n} V_{a_i, k}.
\]

Let $n_e$ and $n_r$ be respectively the numbers of entities and relations in a knowledge graph, $b$ -- a batch size, $d_e$ and $d_r$ -- embedding sizes of the \tuc{} model, $d_h$ -- the size of hyperbolic embeddings. Let the core of the Tucker decomposition be $G \in \mathbb {R}^{d_e \times d_e \times d_r}$, the matrices of embeddings -- the entities embedding matrix $U \in \mathbb {R}^{b \times d_e}$, the relation embedding matrix $T \in \mathbb {R}^{b \times d_r}$ and another entity embedding matrix $V \in \mathbb {R}^{n_e \times d_e}$. 
Similarly, the matrices in the hyperbolic model are $U^{\mathcal{L}} \in \mathbb {R}^{b \times d_h}$, $T^{\mathcal{L}} \in \mathbb {R}^{b \times d_h}$ and $V^{\mathcal{L}} \in \mathbb {R}^{n_e \times d_h}$.

Firstly, we find the complexity of the \tuc{} model. 
To compute the \tuc{} score function, we firstly find $G_1 = G \times_1 U \in \mathbb {R}^{b \times d_e \times d_r}$. 
It takes $O(b  d_e  (d_e d_r))$ operations. 
Then we compute $G_2 = G_1 \times_3 T \in \mathbb {R}^{b \times d_e \times b}$ with $O(b  d_r  (b d_e))$ operations. Finally, the diagonal slice of $G_2$ is taken and multiplied with $V$ with $O(b  d_e  n_e)$ operations. Overall, the total number of operations is:
\[
\begin{aligned}
O_{\textit{TuckER}} = O(b  d_e  (d_e d_r) + b  d_r  (b d_e) + b  d_e  n_e)
= O(b  d_e  (d_ed_r + bd_e + n_e)).
\end{aligned}
\]

Secondly, we find the complexity of the \lp{} model. 
First of all, we compute Lorentz zero element of embeddings $x$  as 
\[
x_0 = \sqrt{\beta + \sum_{i = 1}^d x_i^2},
\]
which is $O(d_h(b + n_e))$ operations. 
Than we compute Lorentz inner products and it takes $O(b d_h n_e + b d_h n_e + b d_h n_e) = O(b d_h n_e)$. Then we compute the sum of all inner products $S_H$  using $O(b n_e)$ operations. Finally, we compute the normalization $N_H$ with $O(b n_e)$ operations and make elementwise division ${S_H}/{N_H}$ with  the complexity $O(b n_e)$. Overall, the \lp{} complexity is:
\[
\begin{aligned}
O_{\textit{TPTF}} = O(d_h(b + n_e) + b d_h n_e + b n_e + b n_e + b n_e) = O(bn_ed_h).
\end{aligned}
\]
The complexity of \mig{} iteration is a sum of \tuc{} and \lp{} complexities and is equal to:
\[
\begin{aligned}
O_{\textit{MIG-TF}} = O_{\textit{TuckER}} + O_{\textit{TPTF}} + O(bn_e) =
 O(b  d_e  (d_ed_r + bd_e + n_e) + bn_ed_h).
\end{aligned}
\]

\slava{In terms of time and parameter complexity, our \mig{} model adds only $30\%$ overhead on training time and $25\%$ overhead on inference time in comparison to \tuc{}.}
    
\section{MIXED GEOMETRY TENSOR FACTORIZATION MODIFICATIONS}
In this section, we describe in more detail additional techniques that we use  for our \mig{} model.

\subsection{Riemannian \mig{}}
To improve the quality of the \tuc{} \citep{TuckER} 
model we use the so-called \emph{RTuckER} model \citep{reim}. This model has the same architecture as \tuc{}, but was trained with the methods of Riemannian optimisation. This Riemannian model significant outperforms \tuc{} on some knowledge graphs. For more details we refer to \citep{reim}. We utilize \emph{RTuckER} as a Euclidean term of our \mig{} model only on WN18RR knowledge graph on which the Riemannian model significantly outperforms \tuc{}.

\subsection{\mig{} with QR decomposition}

\begin{figure}[ht]
	\centering
	\includegraphics[width=1\textwidth]{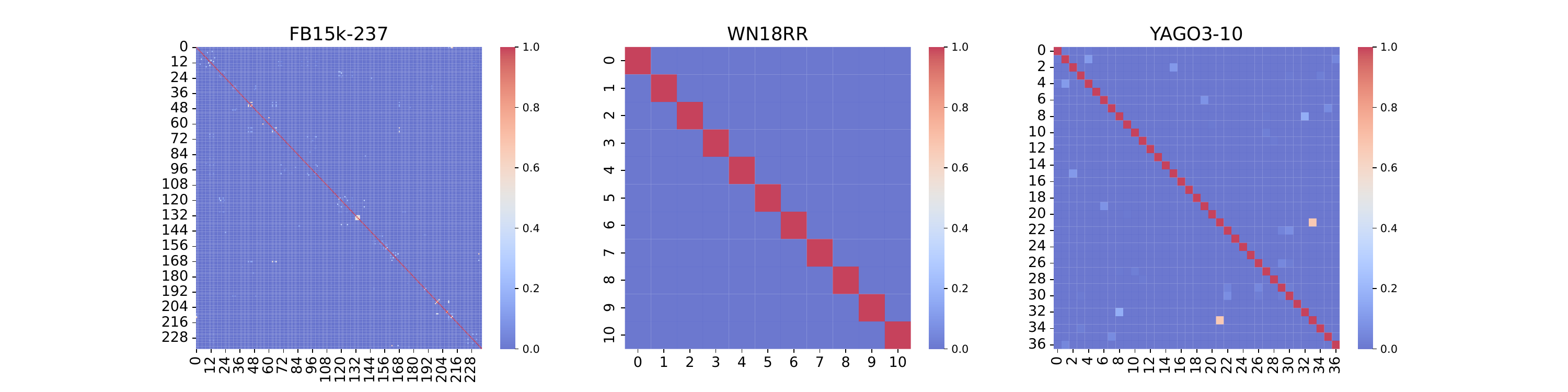}
 
	\caption{Correlation maps for relations of knowledge graphs FB15k-237, WN18RR and YAGO3-10. As can be seen the vast majority of relations in knowledge graphs are uncorrelated. Therefore, the orthoganality of relation embeddings is promising as embeddings of uncorrelated relations do not correlate.}
	\label{fig:heatmaps}
    \end{figure}

 As was shown in Figure~\ref{fig:heatmaps}, most of the relations in knowledge graphs have no correlations, therefore maintaining the orthogonality of relation embeddings creates better vector representation. For this purpose, on each iteration we update the hyperbolic relation embedding matrix $T^{\mathcal{L}}$ with the help of the QR decomposition as follows:
\begin{equation}
T^{\mathcal{L}} = Q^{\mathcal{L}}R. 
\label{eq:qr}
\end{equation}
Then we compute the score function using $Q^{\mathcal{L}}$ instead of $T^{\mathcal{L}}$:

\[
(S_{\textit{MIG-TF}})_i = S_E(G, u, V_i, t) + S_H(u^{\mathcal{L}}, V_i^{\mathcal{L}}, q^{\mathcal{L}}).
\]

In these formulas, $G$ is the core of Tucker decomposition, $u$ and
$u^{\mathcal{L}}$ are the Euclidean and hyperbolic entity embeddings, $V_i$ and
$V_i^{\mathcal{L}}$ are respectively the $i$-th rows of the Euclidean $V$ and of the hyperbolic
$V^{\mathcal{L}}$ entity embedding matrices. The vector $t$ is the
Euclidean relation embedding and $q^{\mathcal{L}}$ is a row of~$Q^{\mathcal{L}}$ from \eqref{eq:qr}.
We call this modified \mig{} as \mig{}$_{QR}$.

The \mig{} model with QR decomposition significantly outperform \mig{}  on knowledge graphs WN18RR and YAGO3-10 where the condition of non correlation of relations is fulfilled almost everywhere. However, \mig{}$_{QR}$ slightly loses to \mig{} on FB15k-237 graph, where the condition of non correlation is usually violated  (see Figure~\ref{fig:heatmaps}).

\section{ROBUSTNESS}
\begin{figure}[ht]
	\centering
	\includegraphics[width=0.6\textwidth]{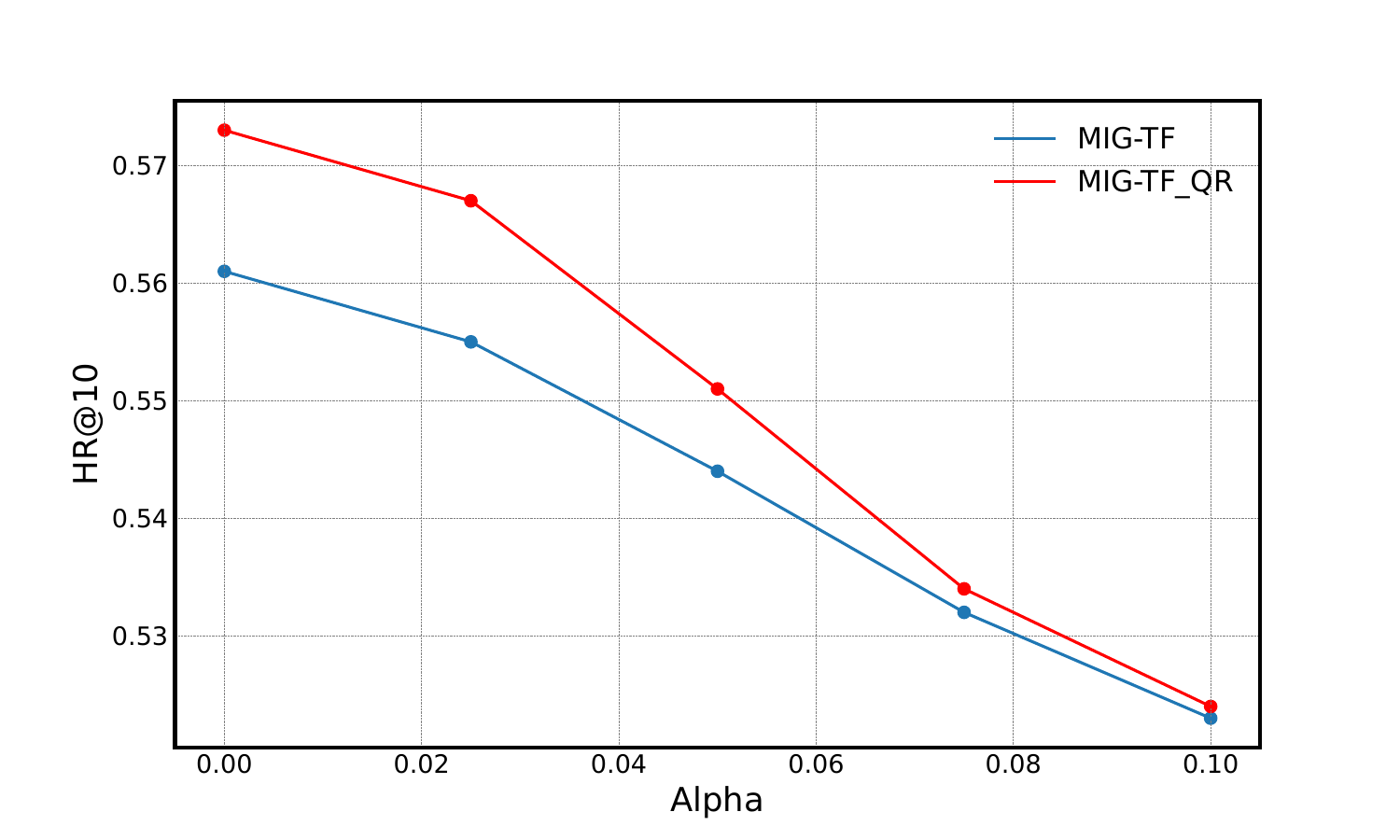}

	\caption{The HR@10 metric on WN18RR knowledge graph depended on the fraction of poison $\alpha$  in training data for $2$ mixed geometry models. }
	\label{fig:p_robustness}
    \end{figure}

To analyze our models' robustness, we add random links in the training data knowledge graph. If $n_{train}$ is the number of links in train data, then we add $\alpha n$ random links to train data. We train the mixed geometry model on poisoned training data and test it on the test data, see Figure \ref{fig:p_robustness}. As we see, even with high levels of noise  ($10\%$), both \mig{} and $\mig_{QR}$ models do not significantly degrade.

\section{THE IMPACT OF EUCLIDEAN AND HYPERBOLIC TERMS}

\begin{figure}[t]
	\centering
	\includegraphics[width=0.6\textwidth]{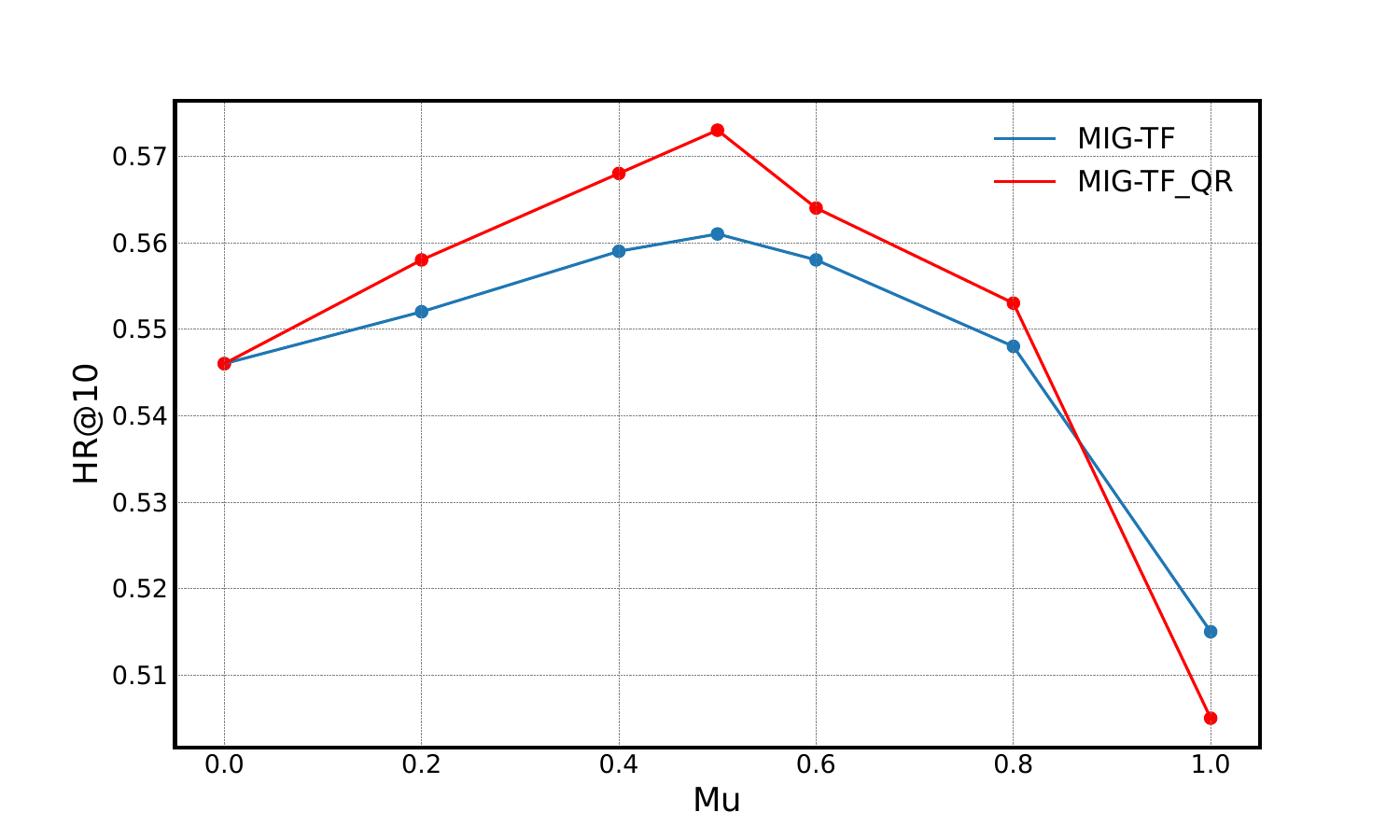}

	\caption{HR@10 against the $\mu$ parameter for $2$ mixed geometry models. The maximum of HR@10 metrics on WN18RR is achieved in $\mu = 0.5$, where the contributions of Euclidean and hyperbolic terms of score function are equal.}
	\label{fig:p_sensitivity}
\end{figure}

In this section, we analyze impact of different terms in the \mig{} score function. 
We represent the score function $S_{\textit{MIG-TF}}$ as:
\[
S_{\textit{MIG-TF}} = \mu S_E + (1 - \mu)S_H, \quad \mu\in(0,1),
\]
where $S_E$ and $S_H$ are respectively Euclidean and hyperbolic components of the score function and $\mu$ is a parameter. By changing $\mu$, we measure the effect of the score function components on the HR@10 metric on WN18RR knowledge graph, see Figure \ref{fig:p_sensitivity}.
It is interesting to note that the best quality is achieved with the equal effect of the Euclidean and hyperbolic terms.
This justifies our choice of the score function.

\section{OPTIMAL CURVATURE}
\begin{figure}[t]
	\centering
	\includegraphics[width=0.6\textwidth]{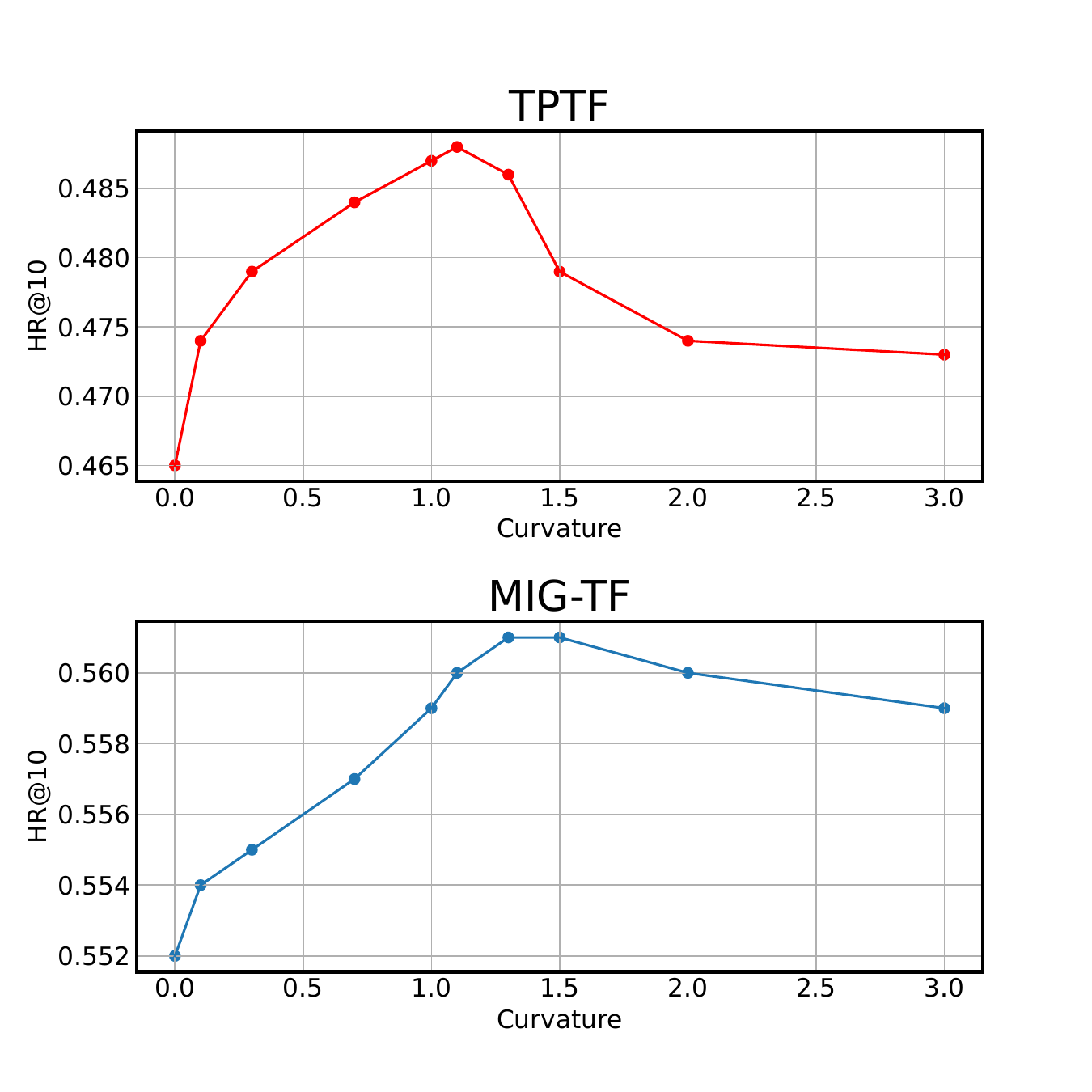}
 
	\caption{The top graph presents the dependence of $HR@10$ metric on WN18RR knowledge graph from curvature of Hyperbolic space $\beta$ for \lp{} model. The bottom one presents the dependence of metric from curvature  for \mig{} model. }
	\label{fig:curvat_2}
    \end{figure}

This section demonstrates dependence of link prediction quality on WN18RR knowledge graph on the curvature in hyperbolic parts of \lp{} and \mig{} models, see Figure \ref{fig:curvat_2}. Note that the optimal curvature for hyperbolic and mixed geometry models are different from each other.

\section{HYPERPARAMETERS}
In this section, we present the hyperparameters of both the Euclidean term (Table~\ref{table:tuc_h}) and the hyperbolic term (Table~\ref{table:mig_h}) of our \mig{} model.
In this tables, $d_e, d_r$ are the sizes of entity and relation embeddings, $lr$ is a learning rate, $\mathcal{N} (0, \rho_e^2I), \mathcal{N} (0, \rho_r^2I)$ -- distributions of initial weights of entities and relations embeddings in the \lp{} component of the hybrid model and  $\beta$ is a curvature of hyperboloid in the Lorentz term of \mig{} model. 
All the experiments we performed were done on one GPU NVIDIA V100.

\begin{table}[ht]
        \caption{Hyperparameters for the \tuc{} component.}
	\label{table:tuc_h}
	\footnotesize
	\centering
	\begin{tabular}{lccc}
		\toprule
  
		$\mathsf{Hyperparametr}$ &  $\mathsf{WN18RR}$ &  $\mathsf{FB15k\text{-}237}$ & $\mathsf{YAGO3\text{-}10}$ \\
		\midrule
            $d_e$ & 200 & 200 & 200 \\
            $d_r$ & 30 & 200 & 30 \\
            $lr$ & 0.01 & 0.001 & 0.003 \\
            $epochs$ & 500 & 500 & 500\\
            $Dropout_1$ & 0.2 & 0.3 & 0.2\\
            $Dropout_2$ & 0.2 & 0.4 & 0.2\\
            $Dropout_3$ & 0.3 & 0.5 & 0.2\\

		\bottomrule
	\end{tabular}
        
\end{table}

\begin{table}[ht]
        \caption{Hyperparameters for the \lp{} component.}
	\label{table:lp_h}
	\footnotesize
	\centering
	\begin{tabular}{lccc}
		\toprule
  
		$\mathsf{Hyperparametr}$ &  $\mathsf{WN18RR}$ &  $\mathsf{FB15k\text{-}237}$ & $\mathsf{YAGO3\text{-}10}$\\
		\midrule
            $d_{\mathcal{L}}$ & 50 & 50 & 50\\
            $lr$ & 0.003 & 0.002 & 0.005 \\
            $\beta$ & 1.3 & 1 & 1.1 \\
            $epochs$ & 250 & 150 & 250 \\
            $\rho_e$ & 0.005 & 0.001 & 0.001 \\
            $\rho_r$ & 0.005 & 0.001 & 0.001 \\
            $Dropout_e$ & 0.2 & 0.3 & 0.2\\
            $Dropout_r$ & 0.2 & 0.3  & 0.2\\
            
		\bottomrule
	\end{tabular}
\end{table}

\begin{table}[ht]
        \caption{Hyperparameters for the \mig{} component.}
	\label{table:mig_h}
	\footnotesize
	\centering
	\begin{tabular}{lccc}
		\toprule
  
		$\mathsf{Hyperparametr}$ &  $\mathsf{WN18RR}$ &  $\mathsf{FB15k\text{-}237}$ & $\mathsf{YAGO3\text{-}10}$\\
		\midrule
            $d_{\mathcal{L}}$ & 50 & 50 & 50\\
            $lr$ & 0.003 & 0.001 & 0.005 \\
            $\beta$ & 1.5 & 1 & 1.1 \\
            $epochs$ & 250 & 100 & 150 \\
            $\rho_e$ & 0.005 & 0.001 & 0.001 \\
            $\rho_r$ & 0.005 & 0.001 & 0.001 \\
            $Dropout_e$ & 0.2 & 0.3 & 0.2\\
            $Dropout_r$ & 0.2 & 0.3  & 0.2\\
            
		\bottomrule
	\end{tabular}
\end{table}


\end{document}